\pdfoutput=1
\documentclass[11pt]{article}

\usepackage[preprint]{acl}
\usepackage{times}
\usepackage{latexsym}

\usepackage[T1]{fontenc}
\usepackage[utf8]{inputenc}

\usepackage{microtype}

\usepackage{inconsolata}

\usepackage{graphicx}

\usepackage{caption}
\usepackage{subcaption}
\usepackage{amsmath}
\usepackage{amssymb}
\usepackage{multirow}
\usepackage{xfrac}
\usepackage{mathtools} 

\usepackage{booktabs}       % professional-quality tables

\usepackage{tablefootnote}

\DeclareMathOperator{\sgn}{sgn}
\newcommand{\abs}[1]{\lvert #1 \rvert}
\newcommand{\PM}{$\pm$}

\title{Robust and Unbounded Length Generalization in Autoregressive Transformer-Based Text-to-Speech}

\author{
  \textbf{Eric Battenberg}\quad 
  \textbf{RJ Skerry-Ryan}\quad
  \textbf{Daisy Stanton}\quad
  \textbf{Soroosh Mariooryad}
\\
  \textbf{Matt Shannon}\quad
  \textbf{Julian Salazar}\quad
  \textbf{David Kao}
\\
  Google DeepMind
\\
  \small{
      \texttt{\{ebattenberg,rjryan,daisy,soroosh,mattshannon,julsal,davidkao\}@google.com}
  }
}

\begin{document}
\maketitle
\begin{abstract}
Autoregressive (AR) Transformer-based sequence models are known to have difficulty generalizing to sequences longer than those seen during training.
When applied to text-to-speech (TTS), these models tend to drop or repeat words or produce erratic output, especially for longer utterances.
In this paper, we introduce enhancements aimed at AR Transformer-based encoder-decoder TTS systems
that address these robustness and length generalization issues.
Our approach uses an alignment mechanism to provide cross-attention operations with relative location information.
The associated alignment position is learned as a latent property of the model via backpropagation and requires no external alignment information during training.
While the approach is tailored to the monotonic nature of TTS input-output alignment, it is still able to benefit from the flexible modeling power of interleaved multi-head self- and cross-attention operations.
A system incorporating these improvements, which we call \textit{Very Attentive Tacotron},
matches the naturalness and expressiveness of a baseline T5-based TTS system,
while eliminating problems with repeated or dropped words and enabling generalization to any practical utterance length.
\end{abstract}

\section{Introduction}

\begin{figure*}[t]
    \includegraphics[width=\columnwidth]{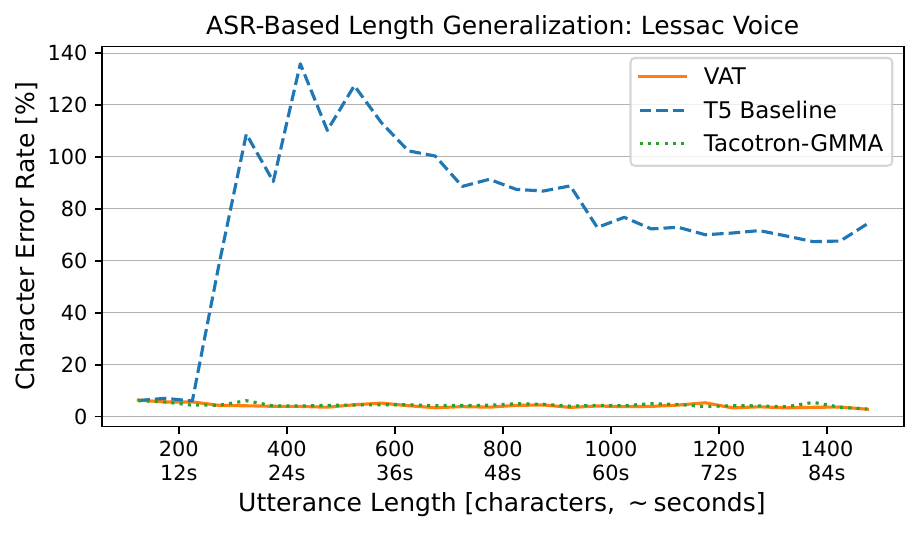}
    \includegraphics[width=\columnwidth]{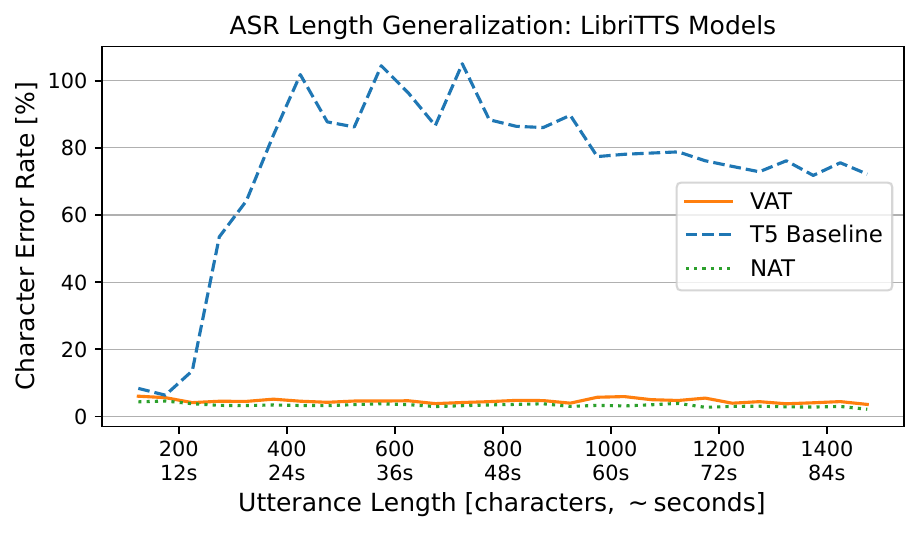}
    \caption{
        %Unlike T5-TTS (the baseline Transformer-based system), Very Attentive Tacotron (VAT) is able to generalize to transcripts of virtually unbounded length despite only training on utterances shorter than 9.6 seconds. 
        Unlike the baseline T5-based TTS system, Very Attentive Tacotron (VAT) is able to generalize to transcripts of virtually unbounded length despite only training on utterances shorter than 9.6 seconds. 
    }
    \label{fig:length-gen}
\end{figure*}

Autoregressive (AR) Transformer-based sequence models \cite{NIPS2017:transformers} are used today in a majority of state-of-the-art systems across text, vision, and audio domains. 
Though swiftly adopted for sequence-to-sequence text-output tasks like machine translation, video captioning \cite{video-captioning-transformer}, and speech recognition \cite{speech-transformer}, adoption for audio-output tasks like text-to-speech (TTS) is notably more recent. 
Inherent difficulties in reliably evaluating TTS systems coupled with the low quality of public TTS datasets likely played a role in the delayed adoption.
This was exacerbated by the failure of AR models of continuous sequences (e.g., spectrograms) to match the performance of their discrete counterparts in the text domain.
It was only after the arrival of audio discretization techniques suitable for speech generation tasks \cite{audiolm} that the TTS community began to direct more effort toward AR Transformer-based models.

A key challenge faced by such models, however, stems from their extensive reliance on attention operations, which tend to cause robustness issues that make them unfit for use in a production TTS environment.
Early attention-based TTS models, such as Tacotron \cite{Wang:2017uz:tacotron,Shen:2017:tacotron2}, often exhibited problems like word omission or repetition, and struggled to generalize 
beyond the training lengths.
Attempts to mitigate these issues using monotonic alignment mechanisms yielded some success 
\cite{Raffel:2017vl:monotonic,Zhang:2018is:forward-attention,battenberg2020:location-relative},
but many ultimately turned to non-autoregressive, duration-based models, which were more robust and efficient during 
synthesis \cite{NEURIPS2020_5c3b99e8:glow-tts,shen2020:non-attentive-tacotron,ren2021fastspeech2}. 
While significant research has focused on improving robustness and length generalization for Transformers in general, these issues still persist in the latest AR TTS systems \citep{song2024ellavstableneuralcodec:ella-v, du2024valltdecoderonlygenerativetransducer:vall-t}.

In this paper we introduce a system called \textit{Very Attentive Tacotron} (VAT), a discrete AR Transformer-based encoder-decoder model designed for robust speech synthesis. 
VAT augments a baseline T5-based \cite{raffel2020exploring:t5} TTS architecture with an alignment mechanism that exploits the monotonic nature of the text-to-speech task, while preserving the powerful modeling capabilities of stacked self- and cross-attention layers.
This leads to virtually limitless length generalization, while matching the naturalness and expressiveness of the T5 baseline system and eliminating issues with repeated or dropped words.

\section{Comparisons to Related Work}
\label{sec:related-work}

Our work was inspired by AudioLM \cite{audiolm}, which showed impressive speech generation results using a decoder-only AR Transformer trained to model discrete targets produced by a self-supervised speech representation and a neural audio codec.
This was extended with text inputs for TTS in the decoder-only case by VALL-E \cite{wang2023neural:arxiv:vall-e}, and in the encoder-decoder case by SPEAR-TTS \cite{kharitonov2023speak:spear-tts} and MQ-TTS \cite{chen2023vector:mq-tts}.

However, \citet{song2024ellavstableneuralcodec:ella-v} found that VALL-E was highly non-robust (in ways we note are reminiscent of early attention-based TTS) 
and proposed ELLA-V, which essentially performs duration modeling by learning to predict ``end-of-phoneme'' markers via forced alignments of training data. 
With similar motivations, VALL-T \cite{du2024valltdecoderonlygenerativetransducer:vall-t} 
extends VALL-E by replacing the objective with a Transducer mechanism \cite{rnn-t} that marginalizes over hard alignments during training and interacts with alignment-shifted text-side relative position embeddings.
Unlike standard AR decoding, this scheme requires a costly sequence-wide inference pass each time the alignment shifts,
and it is unclear whether it can generalize significantly beyond the training lengths.
Finally, to prevent major transcript deviations, MQ-TTS had to limit cross-attention to a single layer and head which is applied over a very narrow (3-6 input tokens) monotonically advancing window.
In summary, current work has not resolved robustness issues without also sacrificing the power and flexibility of the underlying AR Transformer. 

In this work, we focus on encoder-decoder models with cross-attention which we adapt for robustness and length generalization.
The cross-attention operations are informed by a single monotonic alignment position that produces stability without degrading the modeling power of the repeated multi-head cross-attention layers present in the original Transformer architecture. 
The scalar alignment position is learned directly and doesn't rely on dynamic programming or forced alignments during training.

% Very Attentive Tacotron

\section{Very Attentive Tacotron}
\label{sec:very-attentive-tacotron}

\begin{figure*}[t]
  \includegraphics[width=\linewidth]{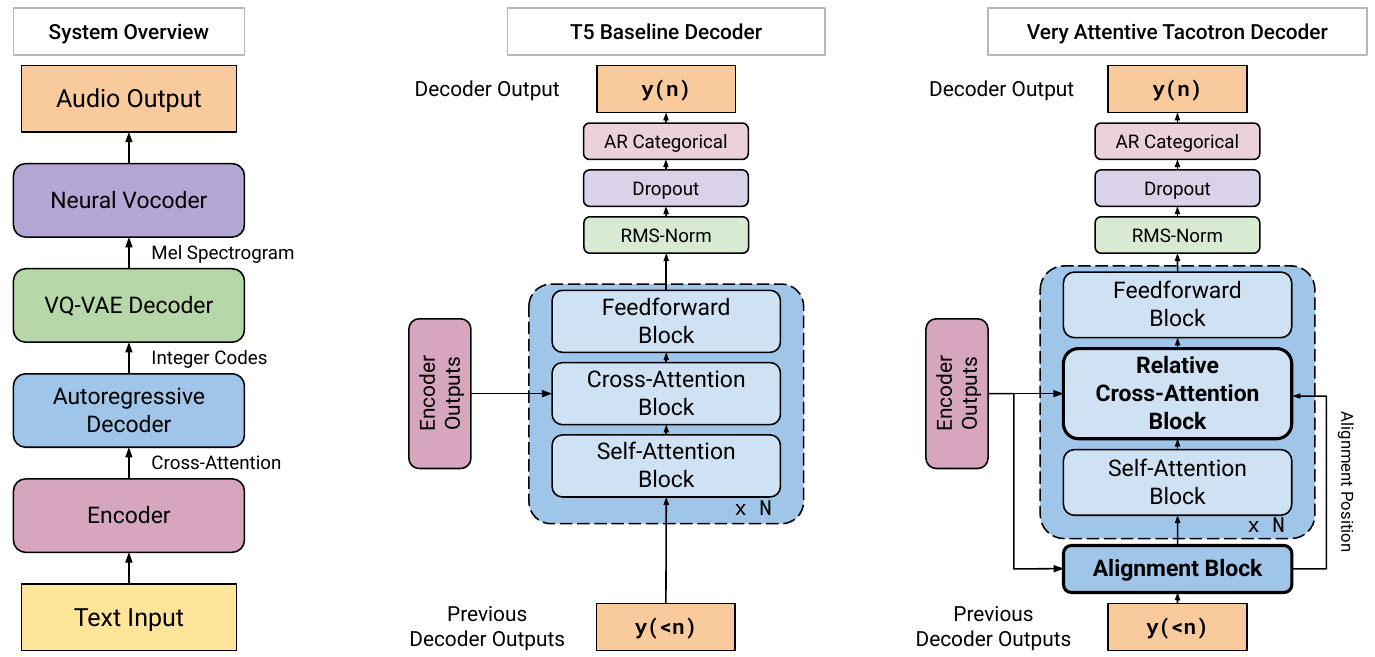}
  \caption{
         High-level discrete AR Transformer TTS system overview (left), T5 baseline decoder based on \citealp{raffel2020exploring:t5} (center), and the VAT decoder (right). Decoder blocks are expanded in Figure~\ref{fig:decoder-blocks}.
 }
  \label{fig:vat-tts-overview}
\end{figure*}

\subsection{System Overview}
\label{subsec:system-overview}

Our VAT model and the baseline T5 model are based on the architecture of the T5 encoder-decoder Transformer originally used for a wide variety of NLP tasks \cite{raffel2020exploring:t5} and later used in SPEAR-TTS \cite{kharitonov2023speak:spear-tts}.
The diagrams in Figure~\ref{fig:vat-tts-overview} give an overview of our discrete TTS setup and then breakdown the differences 
between the decoders used in the two models.

For audio discretization, we use a neural vocoder paired with a VQ-VAE \cite{van2017neural:vqvae} trained to autoencode the vocoder's input spectrograms.
This allows us to quickly retrain the spectrogram discretization for different bitrates or frame rates without having to retrain an entire neural audio codec model, which can be a time consuming and finicky process.
The VQ-VAE is trained using a simple L1 reconstruction loss,
and to improve reconstruction quality, its encoder yields multiple categorical codes per frame using product quantization (PQ) \cite{el-nouby2023image:product-quantization} over multiple codebooks.
The neural vocoder we use is GAN-based and combines features from Parallel WaveGAN \cite{parallel-wavegan} and Hifi-GAN \cite{NEURIPS2020_c5d73680:hifi-gan}.

Unlike the systems covered in Section~\ref{sec:related-work} that do ``zero-shot'' speaker cloning via audio prompting, we learn separate speaker embeddings for each speaker in the training set.
We find this better suited for medium-sized datasets and industry use cases, where we care about creating high quality voices for specific target speakers.

At the input side, we use a self-attention-based encoder to process the text followed by an autoregressive decoder that interacts with the encoded text via multiple cross-attention layers.
We train the decoder to model the sequence of PQ categorical codes produced by the VQ-VAE.
Not only is the decoder trained autoregressively across time, but 
we also model the joint distribution of the PQ codes contained in one frame using an AR decomposition.
Full system details can be found in Appendix~\ref{app:model-details}.

\begin{figure}[tb]
    \begin{subfigure}{\columnwidth}
          \includegraphics[width=\columnwidth]{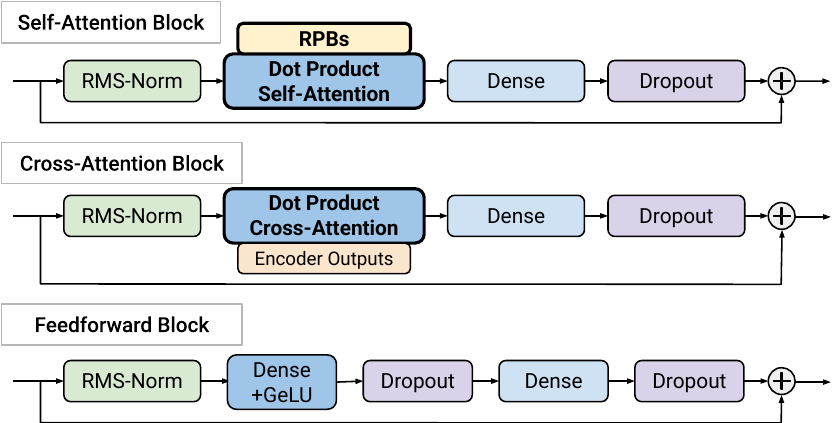}
          \subcaption{Blocks used in the decoder of the T5 baseline architecture.}
          \label{fig:t5-blocks}
    \end{subfigure}
    \begin{subfigure}{\columnwidth}
        \includegraphics[width=\columnwidth]{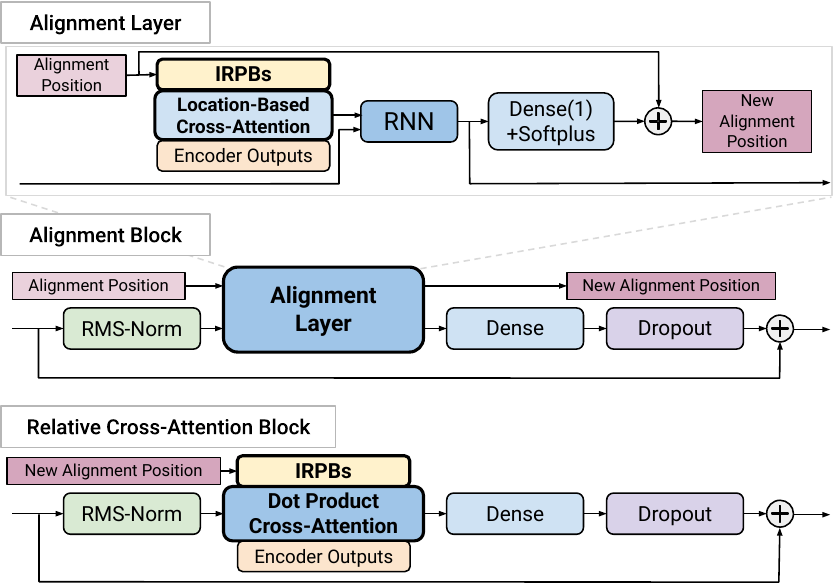}
        \subcaption{New blocks used in the VAT model. The Relative Cross-Attention Block replaces the standard Cross-Attention Block in Figure~\ref{fig:t5-blocks}.}
        \label{fig:vat-blocks}
    \end{subfigure}
    \caption{Diagrams for decoder sub-blocks.}
    \label{fig:decoder-blocks}
\end{figure}

\subsection{T5 Relative Position Biases}

\begin{figure}[t]
  \includegraphics[width=\columnwidth]{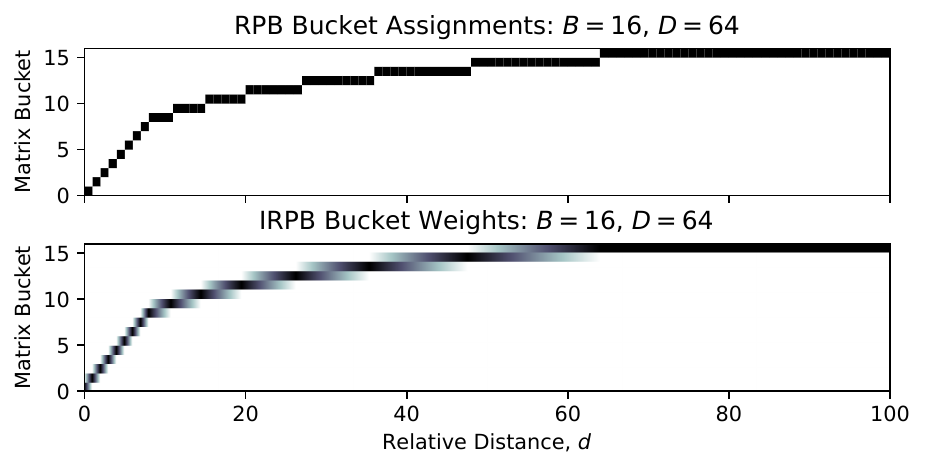}
  \caption{Standard RPB mapping of distances to bias matrix indices (top) and Interpolated RPB mapping of distances to bias matrix index weights (bottom).}
  \label{fig:irpb-mapping}
\end{figure}

T5 introduced an efficient parameterization of relative position biases (RPBs) used to encode locality information in self-attention operations,
which was subsequently shown to outperform other position encoding schemes with respect to length generalization on certain sequence tasks \cite{NEURIPS2023_4e85362c:nope}.
RPBs are used in dot product self-attention when computing attention scores, $s_{i,j}^{(k)}$, and attention weights, $\alpha_{i,j}^{(k)}$, for attention head $k$:
\newcommand{\floorzero}[1]{\lfloor #1 \rfloor_0}
\newcommand{\ceilzero}[1]{\lceil #1 \rceil_0}
\newcommand{\bias}{b_{\floorzero{f(i-j)}}^{(k)}}
\begin{align}
    \label{eq:rpb-attention-scores}
    s_{i,j}^{(k)} &=  
    \frac{\mathbf{q}_i^{(k)} \cdot \mathbf{k}_j^{(k)}}
    {\sqrt{L}}
    + \bias \\
    \label{eq:attention-weights-softmax}
    \alpha_{i,j}^{(k)} &= \frac{\exp\left(s_{i,j}^{(k)}\right)}{\sum_l \exp\left(s_{i,l}^{(k)}\right)}
\end{align} 
where $\mathbf{q}_i^{(k)}$ and $\mathbf{k}_j^{(k)}$ are length-$L$ query and key vectors at positions $i$ and $j$, respectively, and $\floorzero{x} \vcentcolon= \sgn(x)\lfloor\abs{x}\rfloor$ rounds toward zero.
The bias, $\bias$, is taken from a matrix of learned bias values using a function, $f(d)$, to map relative distances into $B$ different buckets:
\newcommand{\hb}{\sfrac{B}{2}}
\begin{multline}
    f(d) = \\
    \begin{cases}
    d,&
    d \in [0,\hb) \\
    \hb + \frac{\log\left(\frac{d}{\hb}\right)}{\log\left(\frac{D}{\hb}\right)}\left(\hb - 1\right),&
    d \in [\hb,D) \\
    B - 1,&
    d \geq D \\
    -f(-d),&
    d < 0 
    \end{cases}
    \label{eq:dist-function}
\end{multline}
So, the first half of the buckets are spaced linearly, the second half logarithmically, and distances beyond $D$ are all mapped to the final bucket.
Negative relative distances are associated with a separate set of $B-1$ buckets which we denote using negative indices.
The top of Figure~\ref{fig:irpb-mapping} shows an example of how distances are mapped to RPB buckets for $B=16$ and $D=64$.

While RPBs provide locality information for self-attention, they can't be used in cross-attention because there is no sense of relative distance between encoder and decoder time steps.
This lack of location information in cross-attention is a big reason why attention-based TTS systems tend to have issues with reliability and length generalization.
One way to introduce a sense of relative distance for use with cross-attention RPBs is to compute an alignment position for each decoder time step that serves as the ``zero'' relative position along the time dimension of the encoder outputs.

\subsection{Interpolated Relative Position Biases}

\newcommand{\fd}{f(d)}
\newcommand{\bi}{\eta}
\newcommand{\fbi}{\lfloor \bi \rfloor}
\newcommand{\fzbi}{\floorzero{\bi}}
\newcommand{\cbi}{\lceil \bi \rceil}
\newcommand{\czbi}{\ceilzero{\bi}}
\newcommand{\ibias}[1]{b_{#1}^{(k)}}

Our approach involves learning the alignment position directly via backprop; however, this means we need to be able to differentiate the RPBs with respect to the alignment position. 
Since standard RPBs only deal with integer relative positions, we cannot do this directly. 
Instead, we bypass the round-toward-zero operation from the bias index expression in eq.~\eqref{eq:rpb-attention-scores} and use $f(d)$ from eq.~\eqref{eq:dist-function} directly as a real-valued bucket index, $\eta$.
This non-integer index is then translated into a bias value by linearly interpolating between the two bias values at the adjacent integer indices in the bias matrix.  
For real-valued bucket index $\bi = \fd$ and bias matrix row $\mathbf{b}^{(k)}$, the interpolated bias for head $k$ and relative distance $d$ can be written as:
\begin{equation}
    \beta^{(k)}(d) = 
    \ibias{\fzbi} + \left(\abs{\bi} - \lfloor\abs{\bi}\rfloor\right)\left(\ibias{\czbi}-\ibias{\fzbi}\right) \\
\end{equation}
where $\ceilzero{x} \vcentcolon= \sgn(x)\lceil\abs{x}\rceil$ rounds away from zero.
The bottom of Figure~\ref{fig:irpb-mapping} shows an example of how relative distances are mapped to the bias index weights used to compute these \textit{interpolated} relative position biases, or IRPBs. 

\subsection{Alignment Layer}
\label{subsec:alignment-layer}

Because we can backprop through IRPBs, the alignment position can be learned directly as a latent property of the model.
We use an RNN-based alignment layer to produce a monotonically advancing alignment position for each decoder time step as shown at the top of Figure~\ref{fig:vat-blocks}.
The RNN is fed with both the input to the alignment layer and the output of a multi-head, location-based cross-attention operation where the attention scores are produced using alignment-informed IRPBs and no content-based query-key comparisons:
\begin{equation}
    s_{i,j}^{(k)} = \beta^{(k)}(p_i-j)
    \label{eq:alignment-attention}
\end{equation} 
where $j$ is the encoder position, $p_i$ is the alignment position at decoder step $i$, and attention weights are produced using the softmax function in eq.~\eqref{eq:attention-weights-softmax}.

%The idea here is that the alignment layer has the fairly simple job of maintaining a rough alignment with the input, while finer-grained, phoneme-level awareness and deeper linguistic understanding can be handled by subsequent cross-attention layers that use content-based queries across multiple heads.
We use purely location-based attention here because the alignment layer has the fairly simple job of maintaining a rough alignment with the input, while finer-grained, phoneme-level awareness and deeper linguistic understanding can be handled by subsequent cross-attention layers that use content-based queries across multiple heads.
This idea is explored further in Appendix~\ref{app:learned-irpbs} where we visualize the learned IRPBs across all layers.

To enforce monotonicity of the alignment, alignment deltas are produced by projecting the RNN output to a scalar and passing it through a softplus function.
We further process the output of the RNN by composing the alignment layer into an alignment block as shown in the center of Figure~\ref{fig:vat-blocks}.

Because the alignment position is unobserved, it cannot be teacher forced, so the alignment layer needs to be executed serially during training;
however, subsequent decoder layers that consume the alignment position can still be run in parallel. 
To minimize the impact of this serialization on training speed, we make the alignment layer as lightweight as possible by using a single-layer LSTM and the location-based attention mechanism in eq.~\eqref{eq:alignment-attention}.

\subsection{Relative Cross-Attention}

To stabilize multi-head cross-attention throughout the rest of the decoder, we augment standard dot product cross-attention with alignment-informed IRPBs:
\begin{equation}
    s_{i,j}^{(k)} = 
    \frac{\mathbf{q}_i^{(k)} \cdot \mathbf{k}_j^{(k)}}
    {\sqrt{L}}
    + \beta^{(k)}(p_i - j) 
    \label{eq:cross-attention-bias}
\end{equation} 
where $\mathbf{q}_i^{(k)}$ is the query at decoder step $i$, $\mathbf{k}_j^{(k)}$ is the key at encoder position $j$,
and attention weights are produced using the softmax function in eq.~\eqref{eq:attention-weights-softmax}.

This operation is used within the relative cross-attention block pictured at the bottom of Figure~\ref{fig:vat-blocks}.
As shown on the right side of Figure~\ref{fig:vat-tts-overview}, the alignment position produced by the alignment layer is fed to every instance of relative cross-attention throughout the model,
and each instance separately attends to the encoder outputs via multi-head relative cross-attention.

\subsection{Initializing IRPBs}
\begin{figure}[tb]
    \includegraphics[width=\columnwidth]{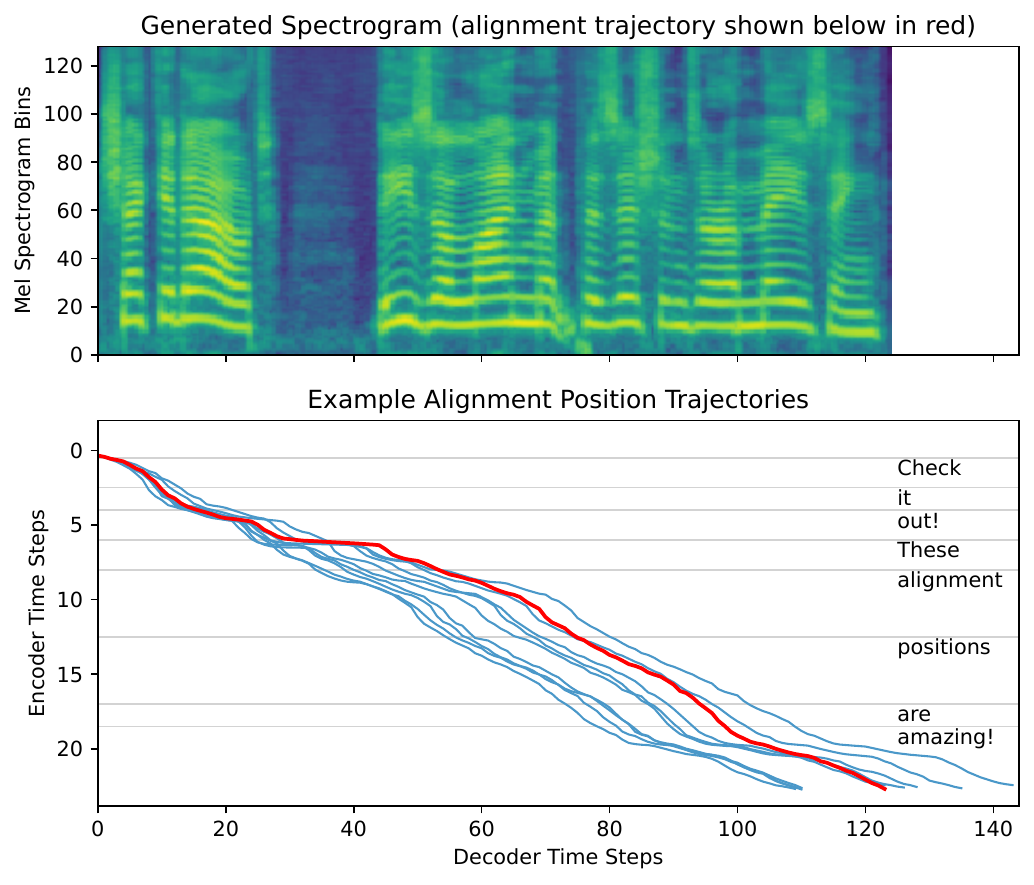}
    \caption{
        Example alignment position trajectories for the transcript 
        ``Check it out! These alignment positions are amazing!'' 
        The trajectory associated with the displayed spectrogram is shown in red.
    }
    \label{fig:alignment-trajectories}
\end{figure}

In order to reliably learn a meaningful emergent alignment position (as visualized in Figure~\ref{fig:alignment-trajectories}), we found it helpful to use a structured initialization scheme for the cross-attention IRPBs.
We chose a Gaussian window centered at zero relative distance with its maximum value normalized to 1.
Due to the softmax operation used in dot product attention, we take the log of the Gaussian window values when initializing the IRPB matrix, but we find it useful to visualize the exponentiated biases in all figures.

\begin{figure}[tbh]
    \begin{subfigure}{\columnwidth}
        \includegraphics[width=\columnwidth]{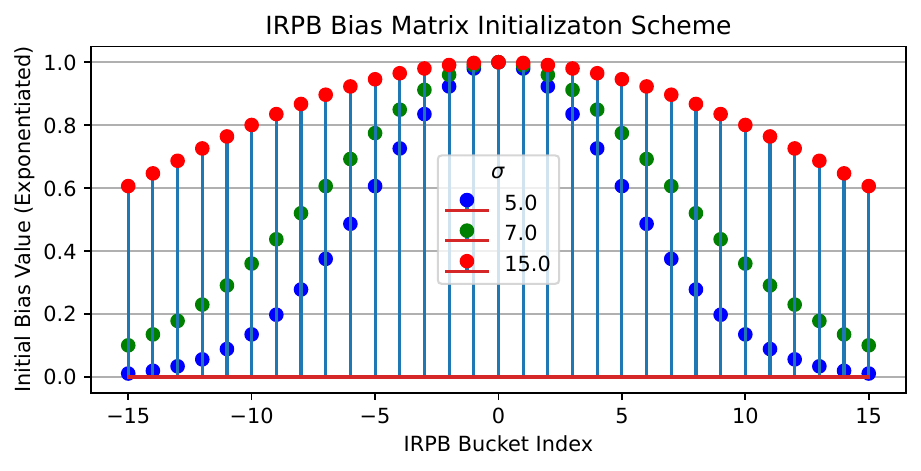}
        \caption{IRPB Gaussian initialization scheme. Note that negative bucket indices denote the biases associated with negative relative distances.}
        \label{fig:irpb-init-scheme}
    \end{subfigure}
    \begin{subfigure}{\columnwidth}
        \includegraphics[width=\columnwidth]{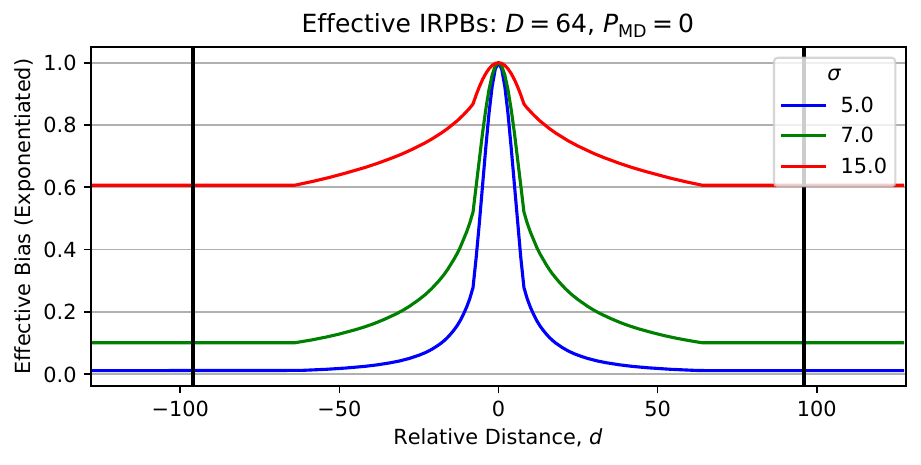}
        \caption{Effective IRPBs using the initialization schemes from Figure~\ref{fig:irpb-init-scheme} with $D=64$ and no MDP. 
        The vertical black lines denote hypothetical maximum training lengths.}
        \label{fig:effective-irpbs-mdp-0}
    \end{subfigure}
    \begin{subfigure}{\columnwidth}
        \includegraphics[width=\columnwidth]{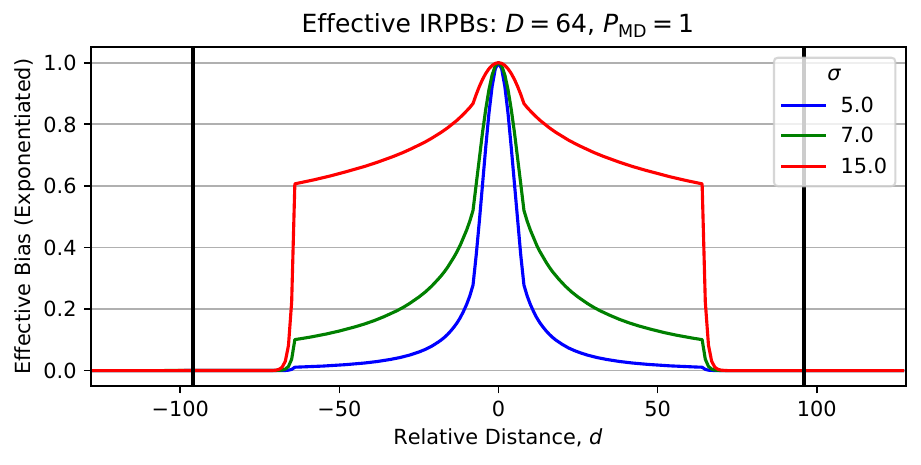}
        \caption{Effective IRPBs using the initialization schemes from Figure~\ref{fig:irpb-init-scheme} with $D=64$ and an MDP of $P_\textrm{MD}=1.0$.}
        \label{fig:effective-irpbs-mdp-1}
    \end{subfigure}
    \caption{Visualizing IRPB matrix initialization.}
    \label{fig:irpb-init}
\end{figure}

Figure~\ref{fig:irpb-init-scheme} shows three examples of this scheme at different standard deviations, $\sigma$.
Given these initial matrix values, the corresponding effective interpolated biases for different relative distances are shown in Figure~\ref{fig:effective-irpbs-mdp-0}.

In early experiments, we found it necessary to initialize the IRPBs using lower standard deviations that heavily suppress attention contributions from relative distances beyond the training lengths.
This was done to prevent undefined behavior at distances not seen during training and was sufficient to guarantee length generalization.

\subsection{Maximum Distance Penalty}
However, using lower standard deviations for IRPB initialization prevents the cross-attention layers from learning longer distance dependencies, which could degrade the model's text understanding capabilities. 
To work around this issue, we incorporate a maximum distance penalty (MDP) that explicitly reduces contributions from relative distances greater than $D$:
\begin{equation}
    \beta_{\textrm{MD}}^{(k)}(d) =
    \begin{cases}
    \beta^{(k)}(d),& 
    \abs{d} < D \\
    \beta^{(k)}(d) - P_\textrm{MD} (\abs{d} - D),& 
    \abs{d} \geq D
    \end{cases}
\end{equation}
where $P_\textrm{MD}$ is the configurable maximum distance penalty.
This allows us to choose wider standard deviations for the Gaussian initialization and still generalize beyond the training lengths.
The effect of this penalty is shown in Figure~\ref{fig:effective-irpbs-mdp-1}.

Despite not being necessary for length generalization in our experiments, we felt it prudent to also apply the MDP to IRPBs used in self-attention layers in order to eliminate additional sources of undefined behavior at relative distances not seen during training.

 % 3

\section{Experiments} % 4

\subsection{Model Configuration}
\label{subsec:model-config}

Our main comparison is between the baseline T5-based TTS model and the augmented VAT model.
We also compare against additional non-Transformer baseline models that were designed for stability, including Tacotron with GMM-based attention (Tacotron-GMMA) \cite{battenberg2020:location-relative} and the unsupervised duration variant of Non-Attentive Tacotron (NAT) \cite{shen2020:non-attentive-tacotron}, a duration-based model.

The reference configuration for the T5 baseline and VAT models uses 6 decoder blocks with 16 attention heads and hidden width 1024. 
For VAT, the alignment block uses a single 256-width LSTM layer and 4 heads in the location-based attention mechanism.
The encoder contains 2 residual convolution stages that downsample the input phoneme sequence by 2x in time, followed by 3 non-causal self-attention blocks with 8 heads and width 512.
All dropout layers use a rate of 0.1.
At test time, we use a sample temperature of 0.7 when sampling from the AR categorical distribution at the decoder output.

The VQ-VAE operates on 80Hz mel spectrograms with 128 bins, downsampling by 2x to produce output at a rate of 40Hz.
Each output frame contains 8 PQ codes with codebook size 256 for an overall bitrate of 2.56 kbps.

For both T5 and VAT, we use position biases with 32 buckets and learn a separate set of biases for every layer, unlike the original T5 paper which shares biases across layers.
For causal self-attention layers in the decoder, all 32 buckets are used for negative relative distances ($B=32$), whereas in cross-attention and non-causal self-attention, the 32 buckets are split evenly between positive and negative distances ($B=16$ per side).
Max distances, $D$, are set to be shorter than the maximum sequence lengths that appear during training,
which are 96 for the encoder outputs (192-length phoneme sequence downsampled by 2x) and 384 for the decoder (9.6-second utterances with 40Hz codes).
We chose max distances of $D=64$ for cross-attention and encoder self-attention and $D=128$ for decoder self-attention.

The T5 baseline uses standard RPBs in all self-attention operations, including in the encoder.
VAT uses IRPBs in all self- and cross-attention operations, including the purely location-based attention in the alignment layer.
IRPBs use an MDP of $P_\textrm{MD} = 1.0$.
Self-attention bias matrices are randomly initialized using a truncated normal, while cross-attention biases use the Gaussian initialization scheme with $\sigma=15$.

Full model configuration details can be found in Appendix~\ref{app:model-details},
and reference implementations for the encoder and decoder are available online.\footnote{\url{https://github.com/google/sequence-layers/blob/main/examples/very_attentive_tacotron.py}}

\subsection{Datasets}
\label{subsec:datasets}
We run experiments using two different English language datasets.
The first is an internal multi-speaker dataset containing a variety of audio, including book-reading, news-reading, and assistant-like utterances.
This dataset consists of 670 hours of audio (\texttildelow700,000 total utterances) spoken by 117 distinct speakers. 
Also included are audiobook recodings from the \emph{Lessac} dataset used in the 2013 Blizzard challenge \cite{lessac-blizzard-2013}, which we use as a common point of comparison with the single-speaker Tacotron-GMMA baseline model. 
The second dataset is the clean-460 subset of LibriTTS \cite{zen2019libritts}, consisting of 213 hours (\texttildelow150,000 utterances) of book-reading audio spoken by 1226 speakers.

\subsection{Training}

We train our T5 and VAT models for 650,000 steps using the Adam optimizer to minimize the negative log-likelihood of the spectrogram VQ codes.
When using the internal multi-speaker dataset, we use the reference configuration from Section~\ref{subsec:model-config}.
When using LibriTTS data, in order to prevent overfitting, we shrink all layer widths to 3/8 scale (e.g., 1024 to 384) compared to the reference configuration.

We use Tacotron-GMMA as a stability-oriented baseline for the internal dataset. 
However, we found that single-speaker versions of the model sounded better than ones trained on the full multi-speaker dataset.
Therefore, we only train Tacotron-GMMA on the Lessac single-speaker data and only evaluate using the Lessac voice when comparing against models trained on the full internal multi-speaker dataset.
    
Non-Attentive Tacotron is used as a LibriTTS baseline and is trained in unsupervised duration mode, following \citet{shen2020:non-attentive-tacotron}.
Full training details are available in Appendix~\ref{app:training-details}.

\subsection{Evaluations}

\paragraph{MOS Naturalness.} We evaluate synthesis quality using a pool of raters to judge naturalness on a 5-point scale. 
For the Lessac voice, we use 885 utterances from the test set, and for LibriTTS we use 900 utterances from the test set. 
We report 99\% confidence intervals along with the mean rating for each model and the ground truth data.

\paragraph{AB7 Side-By-Side (SxS).} Since MOS ratings tend to have calibration and noisiness issues, we complement them with side-by-side naturalness ratings where blinded samples from two models are directly compared on a 7-point ([$-3, 3$]) comparative scale.

\paragraph{ASR-Based Robustness.} For the rated audio, we report character error rate (CER) computed by comparing the input text to the output of a pre-trained speech recognizer that is run on the synthesized audio. 
This addresses the fact that raters don't have access to target transcripts so can't account for dropped or repeated words in their ratings.

\paragraph{ASR-Based Length Generalization.}
Additionally, for each model, we run a length generalization stress test using 1034 transcripts of varying lengths (100--1500 characters) and report CER as utterance length is increased.

\paragraph{Repeated Words Stress Test.} Attention-based TTS models tend to have trouble when repeated words appear in the input text.
We test the ability of the T5 and VAT models to correctly synthesize all the words in three repeated-word templates, each instantiated with 1--9 repetitions of a specific word. 
These templates along with additional evaluation details can be found in Appendix~\ref{app:evaluation-details}.

\section{Results} %5

To aid the reader, audio examples from the naturalness evaluation, length generalization assessment, and repeated words stress test are available online.\footnote{\url{https://google.github.io/tacotron/publications/very_attentive_tacotron/index.html}}

\begin{table}[htb]
    \centering
    \small
    \begin{tabular}{lccr}
    \toprule
    \bf Lessac Voice   & MOS           & SxS vs VAT         & CER\\
    \midrule    
    Ground Truth       & 4.00 \PM 0.07 &                    & 2.9 \\
    VAT                & 3.68 \PM 0.08 & ---                & 3.3 \\
    T5 Baseline        & 3.75 \PM 0.07 & -0.06 \PM 0.14     & 10.2 \\
    Tacotron-GMMA\tablefootnote{Tacotron-GMMA is only trained on the single-speaker Lessac data, whereas the T5 baseline and VAT are trained on the full internal multi-speaker dataset, which includes the Lessac voice.}     
                       & 3.62 \PM 0.08 & -\bf 0.32 \PM 0.14 & 3.7 \\
    \midrule
    \bf LibriTTS       & MOS           & SxS vs VAT         & CER\\
    \midrule
    Ground Truth       & 3.70 \PM 0.09 &                    & 3.6  \\
    VAT                & 3.16 \PM 0.09 & ---                & 4.6  \\
    T5 Baseline        & 3.07 \PM 0.09 & { }0.01 \PM 0.14   & 10.7 \\
    NAT                & 3.22 \PM 0.08 & -0.12 \PM 0.15     & 3.3  \\
    \bottomrule
    \end{tabular}    
    \caption{MOS naturalness, AB7 side-by-side (SxS) versus VAT, and ASR robustness character error rates (CER) for Lessac voice models (top) and LibriTTS models (bottom). Error bars correspond to 99\% confidence intervals.
    }
    \label{tab:results-tables}
\end{table}

\subsection{Naturalness Evaluations}

MOS naturalness results are shown in the left column of Table~\ref{tab:results-tables}.
Note that for both datasets, the confidence intervals are overlapping for all models, so there are no clear winners.
This was surprising given that in informal comparisons the Transformer-based models (T5 and VAT) sounded clearly more expressive and natural to us compared to the Tacotron and NAT baselines.

In cases such as this, side-by-side evals can be helpful due to their increased sensitivity and better calibration.
The naturalness side-by-sides with the VAT model do show that it is preferred over Tacotron-GMMA.
Due to its deterministic regression-based objective, Tacotron sounds less expressive than the Transformer-based models which are trained with a fully probabilistic objective.
VAT also seems to be preferred over NAT, though the result was not quite significant with respect to the 99\% confidence interval.
%Listening to the NAT samples, which sound quite monotonous and robotic, it is clear that its unsupervised duration mechanism was unable to produce a naturally-varying duration predictor. 
Samples from the NAT model sound quite monotonous and robotic, and it is clear that its unsupervised duration mechanism was unable to produce a naturally-varying duration predictor. 
However, some of the raters seemed to prefer its highly enunciated and hyper-intelligible style compared to the more varied and expressive samples produced by VAT and the T5 baseline.
Lastly, VAT is shown to be very even with T5 in terms of naturalness, which is expected given that the two models use very similar architectural backbones and identical training objectives.

\subsection{ASR-Based Robustness}
Despite the similarity in naturalness scores between T5 and VAT, the ASR-based robustness results in Table~\ref{tab:results-tables} show that the T5 baseline produced a significantly higher CER than other models.
This is due to the fact that it tends to drop or repeat words, especially on longer utterances 
(some utterances in the test sets are up to 20 sec long, which exceeds the 9.6 sec training lengths).
Interesting, the NAT model produced a CER below that of the ground truth audio, which is a result of its overly enunciated style.

\subsection{ASR-Based Length Generalization}

Length generalization results are shown in the plots in Figure~\ref{fig:length-gen}
where we plot CER against input text length in characters.
The max training length for the Transformer-based models is 9.6 sec, and we see that soon after the training length is exceeded, the CER of the T5 baseline model sharply increases.
Not only does it drop or repeat individual words, but beyond the training length, it frequently drops or repeats entire clauses and sometimes babbles incomprehensibly.

The other models, including VAT and the two stability-oriented baselines, generalize well all the way up to the max tested lengths (1500 characters, or around 90 seconds). 
We can also see that due to the hyper-intelligibility of the duration-based NAT model, it achieves slightly better CER than the VAT model across all text lengths -- though at the expense of expressivity, as is apparent in the audio examples. 

\subsection{Repeated Words Stress Test}
\label{subsec:repeated-words-results}

The T5 baseline model has difficulty with the repeated words stress test as well.
Over the 27 test phrases, VAT makes no mistakes, whereas T5 makes errors on 14 of the phrases (52\%). 
These errors tend to become increasingly severe as the number of repetitions is increased.
For example, one of the phrases with 9 target repetitions produces 52 repetitions with the T5 model. 
The majority of the mistakes, however, are off by one errors in the number of repetitions, but the T5 baseline produced errors on transcripts that contained as few as 2 repetitions of the target word.
Note that these repetition errors occur even when the synthesis length is shorter than the max training length.

\section{Discussion}

We have shown that the proposed attention enhancements are able to eliminate robustness issues typically observed in Transformer-based TTS systems while matching the synthesis quality of a contemporary T5-based system. 
The VAT model that incorporates these enhancements is able to reliably produce all words in the input text out to seemingly unbounded lengths,
and the alignment-informed, multi-layer, multi-head cross attention it uses is inherently more powerful than the single-phoneme alignment mechanisms used in other robustness-oriented TTS models (see Appendix~\ref{app:learned-irpbs}).

This approach can be directly applied to any encoder-decoder model that uses cross-attention layers in the decoder.
Because the alignment position is constrained to be monotonic, it is best suited for tasks that exhibit broad monotonic alignment between input and output (e.g., TTS and ASR).
However, the flexibility afforded by query-key comparisons within wide IRPB windows should allow the model to adapt well to off-monotonic alignments if needed (e.g., when using unverbalized text). 

% Future work
Due to the scalability of Transformer-based models, future work should test VAT on significantly larger datasets, potentially with encoder or decoder pre-training on unpaired text or audio, respectively. 
Additionally,  using fancier, more powerful approaches to discretization and waveform generation is likely to yield low-level audio quality improvements especially if paired with cleaner, higher-quality datasets.

\section*{Limitations}

\paragraph{Implementation Effort.}
There is additional implementation and configuration effort required for VAT compared to more homogeneous Transformer-based models; 
however, the code and example configurations we provide online should be helpful for researchers attempting to reproduce our work. 

\paragraph{Training Speed Impact.}
The primary practical downside of VAT is the potential for slower training speed due to serialization of the alignment layer during training. 
Because most of the model is still able to be trained in parallel, at the training lengths we use, alignment layer serialization had a relatively small impact on training speed (\texttildelow 12 -- 20\% depending on model size).
Ways to narrow this speed gap further include slimming down the alignment layer RNN or decreasing the VQ-VAE frame rate so that fewer decoder steps are required to model the same amount of audio.

\paragraph{Discrete TTS Comparisons.}
Transformer-based discrete TTS is a rapidly developing area with a short history, so it is difficult to make meaningful comparisons with existing systems, especially since dataset size/quality and model scale can vary drastically. 
Additionally, most discrete TTS systems we encountered are based on prompt-based zero-shot speaker cloning which further complicates direct comparisons. 
Therefore, the discrete TTS baseline we use is based on an existing and well-known Transformer architecture (T5) applied directly to multi-speaker TTS.

\paragraph{Hyper-Parameter Exploration.}
A deeper exploration of the effect of various hyper-parameter choices would be helpful, but was beyond the scope of this initial work.
However, we do cover the most important choices when it comes to enabling robust length generalization in our model.

\paragraph{English Language Focus.}
We use English language datasets in our experiments. Since text-speech alignment tends to be broadly monotonic for the vast majority of written languages, our approach should generalize to other languages; however, this needs to be tested experimentally. 

\paragraph{Potential Risks.}
Our work does not introduce any notable societal or ethical risks beyond those that may already exist for long-form text-to-speech in general.

\bibliography{custom}

\newpage

\appendix

\section{Model Details}
\label{app:model-details}

All models are implemented using TensorFlow.\footnote{\url{https://www.tensorflow.org/}}
The following sections provide additional configuration details.
Example code for the T5 baseline and VAT models is available online.\footnote{\url{https://github.com/google/sequence-layers/blob/main/examples/very_attentive_tacotron.py}}

\subsection{Conv Blocks}
\label{subapp:conv-blocks}
\begin{figure*}[tb]
    \includegraphics[width=\linewidth]{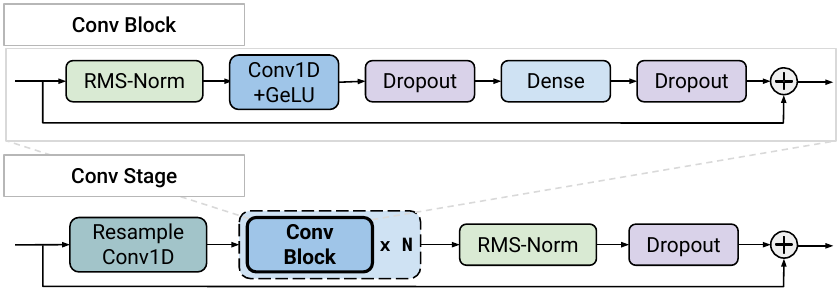}
    \caption{Residual convolution blocks used in text encoder and VQ-VAE.}
    \label{fig:conv-blocks}
\end{figure*}
The residual convolution blocks that we use in the text encoder and VQ-VAE are pictured in Figure~\ref{fig:conv-blocks}.
The basic \textit{Conv Block} uses a 1D conv layer with filter size 3 along with the GeLU nonlinearity. 
A dense layer without a nonlinearity processes the conv output before it is mixed back into the residual line.
A \textit{Conv Stage} composes multiple Conv Blocks after a resampling operation that is implemented using strided convolution for downsampling and transposed convolution for upsampling, which both use filter size 3.
For a specified stage width, $W_\textrm{c}$, we set the number of conv filters and dense units to all be $W_\textrm{c}$ throughout a single Conv Stage.

\subsection{Transformer Blocks}
\label{subapp:transformer-blocks}
The Transformer blocks we use are shown in Figure~\ref{fig:decoder-blocks} and are used in both the text encoder and the decoder.
For each Transformer block, we specify a width, $W$, that is used to set the number of output units in each underlying layer.
For self-attention and cross-attention operations in Figures \ref{fig:t5-blocks} and \ref{fig:vat-blocks}, the dimension used for the queries, keys, and values is simply the block width divided by the number of heads, $W/H$.
All Dense layers have width $W$, with the exception of the ``Dense+GeLU'' layer in the feedforward block, which has a width of $4W$.
The number of units in the RNN (LSTM) in the alignment layer is specified separately.
For the T5 baseline, the self-attention blocks use standard RPBs; and for VAT, self-attention, relative cross-attention, and location-based cross-attention all use IRPBs as described in Section~\ref{subsec:model-config}.

\subsection{Text Encoder}
Here we expand on the encoder description in Section~\ref{subsec:model-config}.
The encoder takes phonemes as input and begins with 2 Conv Stages that each contain 3 Conv Blocks. 
The resampling layer uses a stride of 1 in the first stage and 2 in the second stage for an overall downsampling factor of 2x.

The Conv Stages are followed by 3 Transformer blocks which each contain the self-attention block and feedforward block shown in Figure~\ref{fig:t5-blocks} (unlike the Transformer block used in the decoder, there is no cross-attention).

\newcommand{\We}{W_\textrm{e}}
For a specified encoder width, $\We$, the first Conv Stage has width $\We/2$, while the second has width $\We$.
The widths of the Transformer blocks are $\We$ and attention layers use 8 heads.
For the full-size reference configuration, we use $\We=512$, and for the 3/8 scale LibriTTS configuration, we use $\We=192$.
The example code available online includes text encoder configurations and implementations of all encoder blocks.

\subsection{Decoders}
The T5 and VAT decoder architectures are depicted in Figures \ref{fig:vat-tts-overview} and \ref{fig:decoder-blocks}, and described in Sections \ref{sec:very-attentive-tacotron} and \ref{subsec:model-config}.
Transformer block specifics are covered above in Section~\ref{subapp:transformer-blocks}.
For the full-size reference configuration, we use a decoder width of $W_\textrm{d}=1024$ with 16 attention heads, and for the 3/8 scale LibriTTS configuration, we use $W_\textrm{d}=384$ with 8 attention heads.
In VAT, the alignment layer RNN is an LSTM with width 256 in the reference configuration and 96 in the smaller configuration.
The location-based cross-attention in the alignment layer uses 4 heads.
One element that was not described in the main paper is an initial 1d (causal) convolution with filter size 3 that projects the decoder input (the previous decoder outputs) to the decoder width, $W_\textrm{d}$.
The example code available online includes decoder configurations and implementations of all decoder blocks.

\subsection{VQ-VAE}
The VQ-VAE is covered in the main text in Sections \ref{subsec:system-overview} and \ref{subsec:model-config}.
We implement the VQ-VAE encoder and decoder using a mirrored architecture constructed using the Conv Stages described in Section~\ref{subapp:conv-blocks}.
Both use 2 Conv Stages each containing 4 Conv Blocks. 
The encoder uses widths 512 and 1024 for its 2 stages, while the decoder uses 1024 and 512.
The encoder downsamples by 2x at the beginning of its second stage, and the decoder upsamples by 2x at the beginning of its second stage.

As described in Section~\ref{subsec:system-overview}, the VQ-VAE produces multiple categorical codes per frame using product quantization (PQ) \cite{el-nouby2023image:product-quantization},
To accomplish this, for each frame, the encoder outputs a vector that is divided into $M$ smaller vectors, which are each quantized using separate codebooks.

We use the standard VQ-VAE training objective from \citet{van2017neural:vqvae} with equal weighting between the reconstruction, quantization, and commitment losses. 
The reconstruction term is a simple L1 loss on the 3 second spectrogram segments we use during training.
To improve codebook utilization, we use codebook restarting during training, where individual centroids that drop below a certain moving average usage frequency are reinitialized.

For the spectrogram VQ-VAE models we train on the 2 datasets, we use $M=8$ PQ codebooks which are each comprised of 256 codes.
Because the input spectrograms have a frame rate of 80Hz and the encoder downsamples by 2x in time, this gives an overall quantization bit rate of 2560.

\subsection{Autoregressive Categorical}
Section~\ref{subsec:system-overview} mentions that we model the $M$ categorical codes produced in each VQ-VAE frame
using an autoregressive (AR) decomposition to model the joint distribution of these codes across time.
While the decoder handles the AR decomposition across time, we use a separate AR decomposition to handle the joint distribution of the $M$ codes in a single decoder frame.
Because the number of codes in a single frame is fixed, we can train a separate feedforward network to predict the distribution of each categorical variable given the previous categoricals in the frame and the decoder hidden state:
\newcommand{\ynm}[1]{y_n^{(#1)}}
\begin{multline}
    p(\ynm{m} | \ynm{m-1}, \dots ,\ynm{1}, \mathbf{d}_n) \\
    = f_m(\ynm{m-1}, \ldots ,\ynm{1}, \mathbf{d}_n)
\end{multline}
where $\ynm{m}$ is the $m$th categorical code at frame $n$, $\mathbf{d}_n$ is the decoder hidden state at frame $n$, and $f_m$ is a trainable feedforward network.

These $M$ separate feedforward networks can be trained in parallel using teacher forcing and causal masking of the inputs similar to what is done in Masked Autoregressive Flows \cite{papamakarios2017masked:maf} or MADE \cite{germain2015made:made}.
We implement each function, $f_m$, using 3 dense feedforward layers with the first 2 using the GeLU nonlinearity,
and there is no parameter sharing between functions.
The widths of the layers are set to match the decoder width, $W_\textrm{d}$.

\subsection{Neural Vocoder}
The neural vocoder we use is GAN-based, borrowing from Parallel WaveGAN \cite{parallel-wavegan} and Hifi-GAN \cite{NEURIPS2020_c5d73680:hifi-gan}.
We use the generator from Parallel WaveGAN which consists of 30 layers of dilated convolutions spread over 3 dilation stages. 
The discriminators we use are very similar to the multi-scale and multi-period critics used by Hifi-GAN; 
however, we train both conditional and unconditional versions of them and do not use any feature matching.

\subsection{Parameter Counts}
The full-size reference configurations for T5 and VAT have 138 million and 143 million parameters, respectively. 
The 3/8 scale LibriTTS versions of T5 and VAT have 24 million and 25 million parameters.
The VQ-VAE model we train has 62 million parameters.
The neural vocoder has 39 million parameters.

\section{Datasets Details}
\label{app:datasets}
The LibriTTS dataset \cite{zen2019libritts} is released under a CC BY 4.0 license.\footnote{\url{https://openslr.org/60/}}
The Lessac dataset \cite{lessac-blizzard-2013} that is included in the internal multi-speaker dataset mentioned in Section~\ref{subsec:datasets} contains 150 hours of audio (120,000 utterances) from a single female speaker.
It is released under a non-commercial license as specified at the following URL.\footnote{\url{https://www.cstr.ed.ac.uk/projects/blizzard/2013/lessac_blizzard2013/license.html}}
Our use of these datasets is consistent with their licenses.
  
\section{Training Details}
\label{app:training-details}

\subsection{T5 Baseline and VAT}
The T5 and VAT models were trained for 650k (650,000) steps using the Adam optimizer with $\beta_1=0.9$ and $\beta_2=0.999$, and gradient clipping with threshold 1,000.
The initial learning rate was set using the expression $0.01 / \textrm{sqrt}(\textrm{decoder\_width})$, and the learning rate was decayed to 0.5, 0.25, and 0.1 times the initial learning rate at 500k, 550k, and 600k steps, respectively. 

When training VAT, we initialized the bias in the ``Dense(1)+Softplus'' operation in the Alignment Layer shown in Figure~\ref{fig:vat-blocks} so that it produces initial alignment deltas at about the average rate at which the alignment should proceed across the encoder outputs.  
For a 40Hz VQ-VAE frame rate on the decoder side and 2x downsampled phonemes on the encoder side, the initial bias value was set to $-1.25$, which yields an initial average alignment delta of $0.25$ when passed through the softplus function.
We found that this caused the alignment layer to start producing meaningful alignment trajectories after fewer training steps.

These models were trained using Google Cloud TPUv5e\footnote{\url{https://cloud.google.com/tpu/docs/v5e}}
in a 4x4 topology (16 chips) with a batch size of 128 and maximum audio length of 9.6 seconds.
For the larger reference versions of the models with decoder width 1024, training took about 10 hours for the T5 baseline and 12 hours for VAT.
The smaller 3/8 scale versions with decoder width 384 trained in about 4.2 hours for T5 and 5 hours for VAT. 

\subsection{Spectrogram VQ-VAE}

The spectrogram VQ-VAE models were trained for 1M (1,000,000) steps using the Adam optimizer with $\beta_1=0.9$ and $\beta_2=0.999$ and gradient clipping with threshold 1,000.
The initial learning rate was set to 0.0002 and decayed to 0.5, 0.25, and 0.1 times the initial learning rate at 700k, 800k, and 900k steps, respectively. 
On a 4x4 TPUv5e, these models trained in about 4.6 hours using a batch size of 64.

\subsection{GAN-Based Neural Vocoder}
Our GAN-based neural vocoder was trained for 2M (2,000,000) steps using the Adam optimizer with $\beta_1=0.9$ and $\beta_2=0.999$.
We used initial learning rate 1e-4, halving it at 200k, 400k, 600k, and 800k steps.
Training on a 4x4 TPUv5e took about 33 hours using a batch size of 64 and audio segment size of 300ms.

\subsection{Tacotron-GMMA}
We trained the Tacotron-GMMA model on the Lessac dataset, following the training procedure for the GMMv2b model from \citet{battenberg2020:location-relative}.
This involved training for 300k (300,000) steps using the Adam optimizer and a gradient clipping threshold of 5.
The learning rate was initially set to 1e-3 and reduced to 5e-4, 3e-4, and 1e-4, at 50k, 100k, and 200k steps, respectively.
Training on a 4x4 TPUv5e took around 8 hours using a batch size of 256 and maximum audio length of 5 seconds.

The associated WaveRNN vocoder was trained for 2.5M (2,500,000) steps using the Adam optimizer.
Teacher forced predicted spectrograms produced by the Tacotron model were used as input to the vocoder during training, rather than ground truth spectrograms.
The initial learning rate was set to 2e-4 and then decayed to 1e-4, 5e-5, 2e-5, 1e-5, and 5e-6 at 250k, 450k, 650k, 850k, and 1M steps, respectively. 
Training took around 72 hours (3 days) using a batch size of 128 and waveform segment size of 37.5ms.

\subsection{Non-Attentive Tacotron}
We trained the NAT model using the unsupervised duration training procedure from \citet{shen2020:non-attentive-tacotron}.
Training was done using Google Cloud TPUv3\footnote{\url{https://cloud.google.com/tpu/docs/v3}}
in a 4x4 configuration (16 chips).
Training the NAT model for 150k (150,000) steps took 91.5 hours (3.8 days).
Pre-training the WaveRNN vocoder for 500k steps took 74.5 hours (3 days). 
And the joint fine-tuning of both for 500k steps took 146 hours (6 days).

\section{Evaluation Details}
\label{app:evaluation-details}

\subsection{MOS and Side-By-Side Ratings}
To gather the MOS and side-by-side naturalness ratings, we used a pool of professional raters recruited for their ability to judge the quality of English speech. 
Raters are paid per rating at a rate dependent on average task length. 
Individual raters were limited to 6 ratings for a single evaluation experiment, and each item was rated once.
For both tasks, we report mean scores along with 99\% confidence intervals.\footnote{\url{https://www.itl.nist.gov/div898/handbook/eda/section3/eda352.htm}}

Rating templates were reviewed and approved by an internal group responsible 
for wording clarity and overall rater experience.

\paragraph{MOS Naturalness}
For MOS naturalness, the rater instructions were as follows: 
\begin{quote}
In this task, you will be given one or more audio clips. For each clip, please listen to the speech very carefully and then select a rating for each audio clip. The rating should be based on how natural or unnatural the sentence sounded. Please do not judge the grammar or the content of the sentence. Instead, just focus on how natural the speech sounds.
\end{quote}
The rater then answers the question ``How natural does the speech sound?'' by choosing a response on a 5-point scale with labels: Bad, Poor, Fair, Good, Excellent.

\paragraph {Side-by-Side Naturalness}
For side-by-side naturalness, the rater instructions were as follows: 
\begin{quote}
In this task, your job is to listen to two different audio samples containing speech. The text spoken will be the same for both Speech Samples. Please listen to both samples before selecting a rating.
\end{quote}
The rater then answers the question ``Which side sounds better?'' using the horizontally arranged choices: Much Better, Better, Slightly Better, About The Same, Slightly Better, Better, Much Better; which correspond to integers in [-3, 3] when tallying scores.

\subsection{ASR Length Generalization}
The 1034 transcripts we use in the length generalization metric are between 100 and 1500 characters long 
and were taken from the first Harry Potter novel. 
The longest transcripts tend to yield speech up to around 90 seconds in length.

\subsection{Repeated Words Stress Test Templates}
The 3 repeated word templates we used in the repeated words stress test are shown below. 
For each template, the word that is repeated between 1 and 9 times is shown in square brackets. 
In the transcripts, repetitions are joined using a comma followed by a space.
The templates are listed below:
\begin{enumerate}
    \item I am [really], super duper tired.
    \item My phone number is 1, 800, [9], 2.
    \item Wow! That's [pretty] good!
\end{enumerate}

\section{Visualizing Learned IRPBs}
\label{app:learned-irpbs}

\subsection{IRPB Visualizations}
In Section~\ref{subsec:alignment-layer}, we mentioned that the alignment layer was designed to maintain a rough alignment with the input, and therefore, it wasn't crucial for it to handle exact phoneme-level alignment due to the flexibility afforded by the content-based query-key comparisons in multi-head relative cross-attention.

In this section, we show the learned IRPB values from the larger reference version of the VAT model when trained on the internal multi-speaker dataset. 
In the figures, all bias values are exponentiated to aid visual comparisons and to account for their use inside of a softmax when computing attention weights.
On the left side of each figure, we show the biases for each head as overlapping scalar plots, with the lines colored according to their peakedness (kurtosis).
On the right we show the same values in image format where each row contains the biases for a single head, and the rows are sorted according to their peakedness.

The x-axis denotes the index in the bias matrix, where negative indices are associated with negative relative distances.
Note that encoder self-attention, alignment cross-attention, and relative cross-attention use both positive and negative relative distances, whereas decoder self-attention only uses negative relative distances because it is causal. 

\subsection{Encoder Self-Attention IRPBs}
\begin{figure*}[p]
    \includegraphics[width=\linewidth]{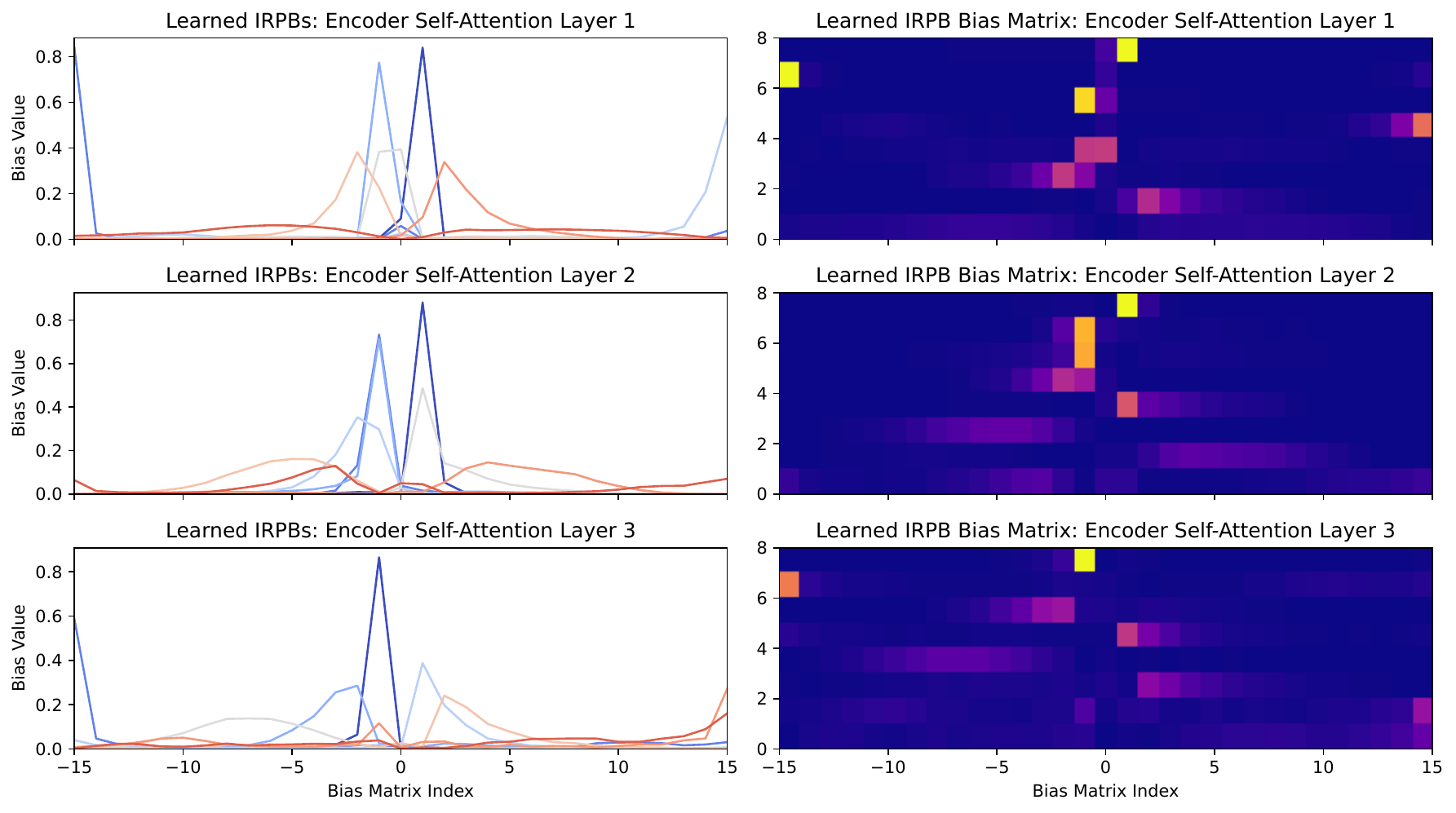}
    \caption{Learned IRPBs for encoder self-attention layers.}
    \label{fig:encoder-self-learned-irbps}
\end{figure*}

In Figure~\ref{fig:encoder-self-learned-irbps}, we see the IRPBs learned in each layer of the (non-causal) self attention layers in the text encoder.
We can see that the 8 heads are arranged in an interesting lobed pattern at a variety of scales, presumably to handle things like syllable formation and different types of text understanding.  
There are also heads that focus on the edges of the bias index range, which corresponds to a max distance of $D=64$.

\subsection{Alignment Layer Location-Based Cross-Attention IRPBs}
\begin{figure*}[p]
    \includegraphics[width=\linewidth]{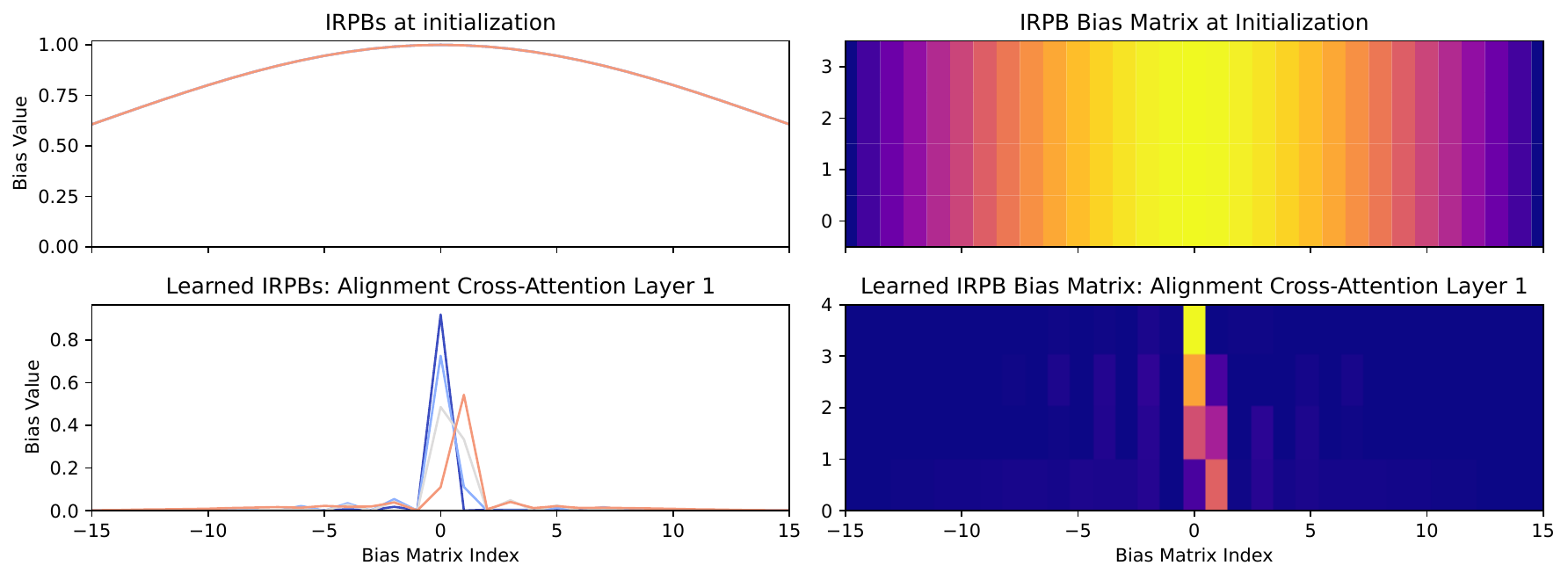}
    \caption{Learned IRPBs for alignment layer location-based cross-attention. Also shown are the initial IRPB values when using the Gaussian initialization scheme with $\sigma=15$ (top).}
    \label{fig:alignment-cross-learned-irbps}
\end{figure*}

IRPBs for the location-based cross-attention operation contained within the alignment layer (eq.~\eqref{eq:alignment-attention}) are shown in Figure~\ref{fig:alignment-cross-learned-irbps}.
At the top, we also show the IRPBs at initialization (when using the Gaussian initialization scheme with $\sigma=15$).
Because this purely location-based mechanism doesn't use content-based query-key comparisons, the 4 heads learn very simple IRPBs that distribute their weight mainly between the current alignment position (relative distance 0) and one step ahead.
This is likely because the primary task of the alignment layers is to decide whether to move the alignment position forward. 
Deeper understanding of the text and phonetics can be handled by subsequent relative cross-attention layers.

\subsection{Relative Cross-Attention IRPBs}
\begin{figure*}[p]
    \includegraphics[width=\linewidth]{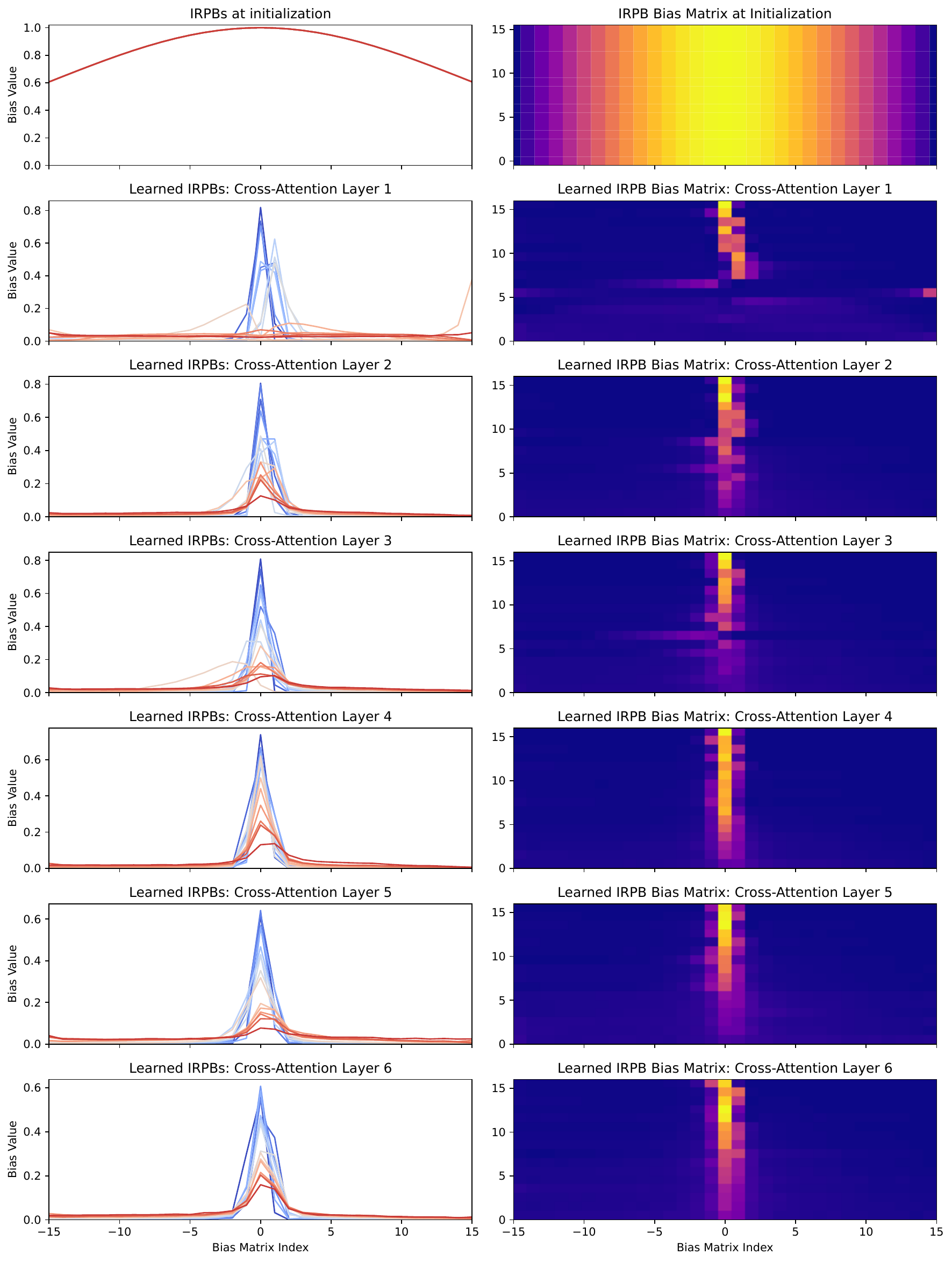}
    \caption{Learned IRPBs for relative cross-attention layers.
    Also shown are the initial IRPB values when using the Gaussian initialization scheme with $\sigma=15$ (top).}
    \label{fig:cross-learned-irbps}
\end{figure*}

Figure~\ref{fig:cross-learned-irbps} shows the IRPBs learned for the 16 heads of each of the 6 relative cross-attention layers (using the expression in eq.~\eqref{eq:cross-attention-bias}). 
Also shown at the top are the IRPB values at intialization.
Across the 6 layers, we see that much of the IRPB weight is centered at relative distance zero (the current alignment position) but each head seems to learn a different amount of focus, with some learning a quite broad focus that can even reach out to the max distance, $D=64$.
We also see that some of the layers (especially the first layer), contain heads with learned bias windows that are offset from the zero position or have a lobed configuration that looks forward or backward in time. 
The fact that the IRPBs used in relative cross-attention layers haven't all just collapsed to focus solely on the current alignment position suggests that the model is able to benefit from the ability to cross-attend to the input in more complicated ways. 

\subsection{Decoder Self-Attention IRPBs}
\begin{figure*}[p]
    \includegraphics[width=\linewidth]{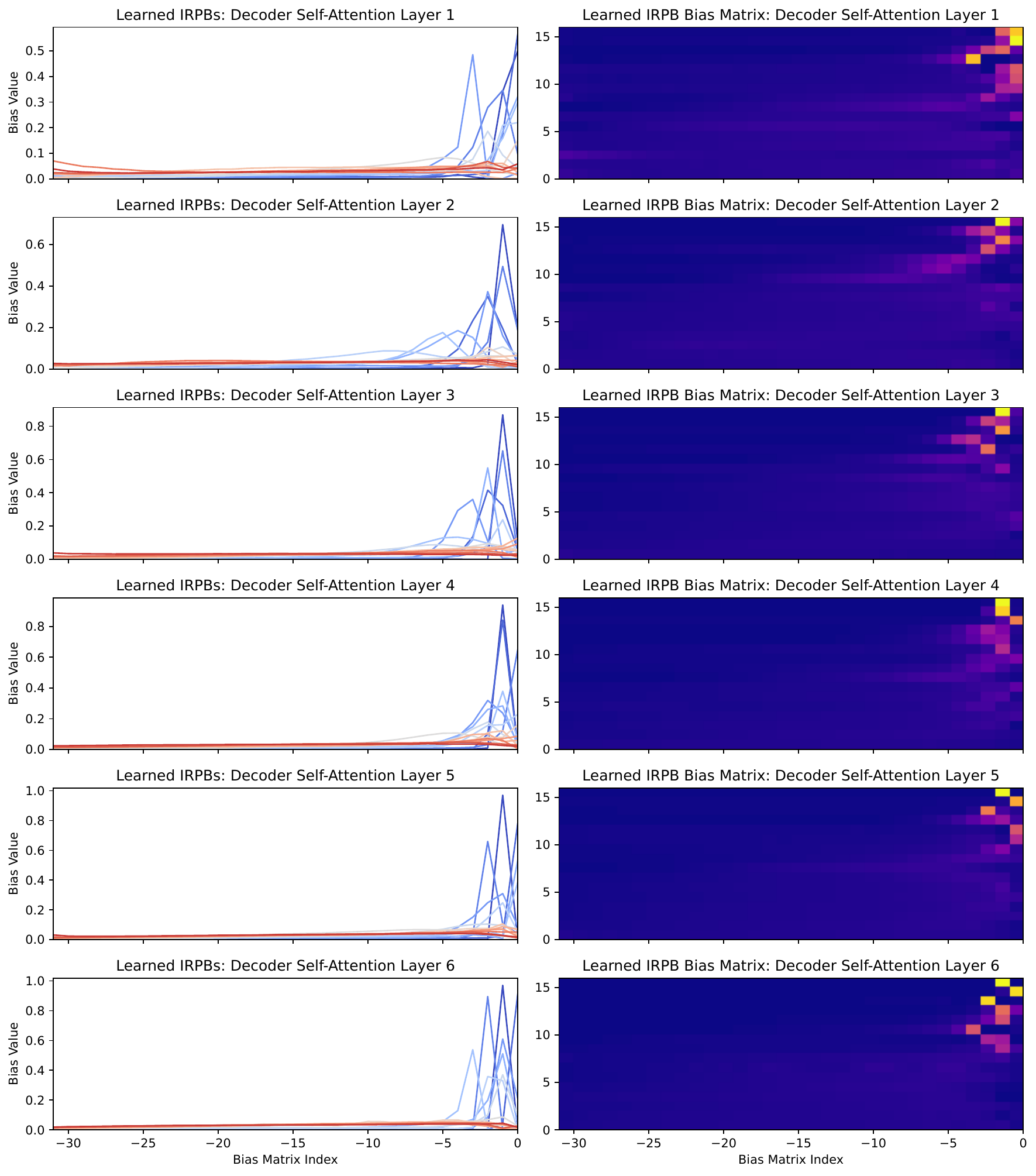}
    \caption{Learned IRPBs for decoder self-attention layers (which use causal self-attention, so only negative relative distances are used).}
    \label{fig:decoder-self-learned-irbps}
\end{figure*}

Lastly, Figure~\ref{fig:decoder-self-learned-irbps} shows the IRPBs learned by the each of 6 decoder self-attention layers across their 16 heads.
Because these layers use causal self-attention, all 32 bias buckets for each head are used for negative relative distances. 
The IRPB windows appear to be arranged so that some of the heads focus on the recent past, while others have a very broad focus that reaches all the way out to the max distance, $D=128$.

\end{document}